\documentclass[11pt]{article}

\usepackage[final]{acl}

\usepackage{times}
\usepackage{latexsym}
\usepackage{dblfloatfix}
\usepackage{placeins}

\usepackage[T1]{fontenc}
\usepackage{amsfonts}
\usepackage[utf8]{inputenc}

\usepackage{microtype}

\usepackage{inconsolata}

\usepackage{graphicx}
\usepackage{booktabs}

\usepackage{graphicx}
\usepackage{subcaption}
\title{Do Personality Traits Interfere? Geometric Limitations of Steering in Large Language Models }

\author{
 \textbf{Pranav Bhandari\textsuperscript{1,*}}, 
  \textbf{Usman Naseem\textsuperscript{2}},
 \textbf{Mehwish Nasim\textsuperscript{1,*}}
\\
 \textsuperscript{1}Network Analysis and Social Influence Modelling (NASIM) Lab \\
 School of Physics Maths and Computing\\
 The University of Western Australia\\
 \textsuperscript{2}School of Computing, Macquarie University
\\
  \small{
   \textbf{\textsuperscript{*}Correspondence:} {firstname.lastname@uwa.edu.au}
 }
}

\begin{document}
\maketitle
\begin{abstract}
Personality steering in large language models (LLMs) commonly relies on injecting trait-specific steering vectors, implicitly assuming that personality traits can be controlled independently. In this work, we examine whether this assumption holds by analysing the geometric relationships between Big Five personality steering directions. We study steering vectors extracted from two model families (LLaMA-3-8B and Mistral-8B) and apply a range of geometric conditioning schemes, from unconstrained directions to soft and hard orthonormalisation. Our results show that personality steering directions exhibit substantial geometric dependence: steering one trait consistently induces changes in others, even when linear overlap is explicitly removed. While hard orthonormalisation enforces geometric independence, it does not eliminate cross-trait behavioural effects and can reduce steering strength. These findings suggest that personality traits in LLMs occupy a slightly coupled subspace, limiting fully independent trait control.
\end{abstract}

\section{Introduction and Background}
%%first para -- points can be ai, psych, advancements, problem
Large Language Models (LLMs) have demonstrated significant advances in their ability to exhibit personality traits \cite{jiang2024personallm, jiang2023evaluating, serapio2023personality}, often aligned with the Big Five personality framework. Extensive prior work has explored the evaluation \cite{bhandari2025evaluating, pellert2024ai}, extraction \cite{jiang2024personallm}, and steering \cite{zhu2024personality,chen2025persona,bhandari2025activation} of personality traits in the literature. Recent advances in activation engineering have enabled the steering of LLM behaviour by injecting \emph{activation vectors} into the model’s residual stream to control the strength of trait expression. This approach offers several advantages over fine-tuning \cite{zhu2024personality}, which is typically a heavy and resource-intensive process. Model behaviour can instead be adjusted at inference time using a \emph{precise and controllable “knob”}, allowing for flexible and efficient personality modulation.

However, these studies do not emphasise the inherently entangled nature of personality traits, commonly referred to as the OCEAN traits (Openness, Conscientiousness, Extraversion, Agreeableness, and Neuroticism), and instead draw conclusions based on the independent manipulation of individual traits. Our findings indicate that models do not learn human psychological constructs as orthogonal basis vectors; rather, they acquire entangled representations shaped by the correlations present in their training data. Empirically, we demonstrate that increasing a single dimension, such as Openness, does not result in an isolated change along that axis. Instead, it simultaneously increases other dimensions, including Agreeableness, Conscientiousness, and Extraversion, while decreasing Neuroticism. These observations suggest that, rather than reflecting shifts along a single trait dimension, such changes are better understood as movements along a broader social axis.

Previous work has shown that high-level behaviours in large language models can often be approximated as linear directions in the activation space, enabling behaviour steering via difference-of-means or PCA-based vectors \cite{zou2023representation}. However, such methods typically assume independence between concept directions. The concept-erasure literature studies how correlated information can be removed from representations, such as with Iterative Nullspace Projection \cite{ravfogel2020null}, introducing greedy, order-dependent projections, and LEACE: Perfect Linear Concept Erasure in Closed Form \cite{belrose2023leace}, proposing a global, order-independent solution. Also, interpretability work such as Toy Models of Superposition \cite{elhage2022toy} shows that models often encode correlated concepts in superposition, suggesting that geometric orthogonality alone may not guarantee semantic independence. Our work builds on these insights by applying both greedy and global Orthogonalisation strategies to personality steering and empirically evaluating their effect on cross-trait interference.
% \section{Literature Review}

The overarching aim of this work is to systematically understand the personality trait vectors used in activation engineering. Prior activation engineering methods typically apply one trait direction at a time and evaluate whether the target trait shifts as intended \cite{sun-etal-2025-personality,yang2024exploring,bhandari2025activation}, while leaving the effects on non-target traits unexamined. We systematically analyse the geometric relationships between Big Five personality directions and test how different constraint strategies, ranging from no constraints to full orthonormalisation, change both trait specificity and downstream behaviour. Rather than assuming that traits should be fully separable, we treat interference as an empirical phenomenon to be measured. Our goal is to understand when enforcing geometric independence helps, when it harms steering, and what this reveals about how personality traits are encoded inside the model.

We propose the following research questions: 
\begin{enumerate}
    \item \textbf{RQ1:} We hypothesise that personality steering directions are geometrically independent in large language models.
    \item \textbf{RQ2:} What happens to steering effectiveness when geometric independence is explicitly enforced?
    \item \textbf{RQ3:} Are observed trait dependencies consistent across different model families?
\end{enumerate}

\section{Methodology}
The scope of this work focusses on understanding the behaviour of personality steering vectors and their interactions. We adopt the hybrid layer selection approach from \cite{bhandari2025activation} as our steering baseline, due to its effectiveness while preserving the model capacity. Our methodology then builds on this framework to analyse geometric dependencies between personality traits under different constraint settings.

\paragraph{Steering Mechanism.}
%% repeated from last para

% We adopt the hybrid layer-based steering framework \cite{bhandari2025activation} to control personality traits at inference time. 

Trait directions are estimated from high/low labelled data \cite{li2025big5} at each layer and aggregated into a single weighted direction per trait, capturing consistent personality signals across the network. An offline prior layer is selected for each trait using neutral probe prompts by applying a small steering signal and measuring distributional sensitivity at the next token. At runtime, a lightweight dynamic check adapts the layer choice to the input prompt. Steering is applied via forward hooks using projected and intensity-scaled trait vectors, and evaluated under base, positive, and negative settings using both personality questionnaires and generation-based assessments.

\paragraph{Trait Direction Conditioning (C0--C5).}
Let $\mathbf{d}_c \in \mathbb{R}^D$ denote the normalised weighted steering direction for trait $c$. Empirically, these directions are not independent and exhibit substantial cosine overlap. To study how geometric constraints affect steering behaviour and cross-trait interference, we construct the following conditioning schemes:

\begin{itemize}
  \item \textbf{C0 (Baseline).}  
  Original trait directions $\mathbf{d}_c$ are used without modification.

  \item \textbf{C1 (Soft Symmetric Whitening).}  Directions are stacked into $\mathbf{D}$ and transformed via a regularised Gram matrix: $\mathbf{D}'=((1-\gamma)\mathbf{G} + \gamma\mathbf{I})^{-1/2}\mathbf{D}$, where $\gamma \in (0,1)$ is a shrinkage parameter. This scales down off-diagonal correlations without forcing strict orthogonality, preserving more of the original shared geometry than hard whitening.

  \item \textbf{C2 (Greedy Orthogonalisation).}  
  A Gram--Schmidt procedure sequentially removes projections $\langle \mathbf{d}_i,\mathbf{d}_j\rangle\mathbf{d}_j$, yielding an orthonormal basis that is order-dependent.

  \item \textbf{C3 (Selective Orthogonalisation).}  
  Projection is applied only when $|\cos(\mathbf{d}_i,\mathbf{d}_j)|>\tau$, preventing over-disentanglement while suppressing dominant overlaps.

  \item \textbf{C4 (Soft Projection).}  
  Correlated components are partially attenuated as $\mathbf{d}_i \leftarrow \mathbf{d}_i - \beta\langle \mathbf{d}_i,\mathbf{d}_j\rangle\mathbf{d}_j$ when $|\cos|>\tau$, trading off disentanglement strength and retention.

  \item \textbf{C5 (Hard Orthonormalisation).}  
  A symmetric Löwdin transformation enforces $\mathbf{D}'\mathbf{D}'^\top=\mathbf{I}$, completely removing linear overlap in an order-independent manner.
\end{itemize}

All conditions use the same steering injection and hybrid layer selection mechanism, isolating the impact of geometric constraints on steering efficacy and trait interference.

\begin{table*}[t]
\centering
\small 
\setlength{\tabcolsep}{5pt} 
\renewcommand{\arraystretch}{1.25} % 
% \caption{\textbf{Trait-Level Steering Contrast under Geometric Constraints.}
% Comparison of target steering contrast ($T$, Intended Target) and maximum cross-trait contrast ($B_{\max}$, Unintended Target) for LLaMA-3-8B and Mistral-8B under baseline (C0), soft-constrained (C4), and hard orthonormal (C5) trait vector constructions. Values report the difference between positively and negatively steered generations (High$-$Low) as measured by judge scores (1--5 scale). Parentheses indicate the trait responsible for $B_{\max}$.}

\resizebox{\textwidth}{!}{%

\begin{tabular}{l | cc cc cc | cc cc cc}
\toprule
& \multicolumn{6}{c|}{\textbf{Llama-3-8B}} & \multicolumn{6}{c}{\textbf{Mistral-8B}} \\
\cmidrule(lr){2-7} \cmidrule(lr){8-13}
& \multicolumn{2}{c}{\textbf{C0 (Base)}} & \multicolumn{2}{c}{\textbf{C4 (Soft)}} & \multicolumn{2}{c|}{\textbf{C5 (Hard)}} 
& \multicolumn{2}{c}{\textbf{C0 (Base)}} & \multicolumn{2}{c}{\textbf{C4 (Soft)}} & \multicolumn{2}{c}{\textbf{C5 (Hard)}} \\
\textbf{Target Trait} & $T$ & $B_{max}$ & $T$ & $B_{max}$ & $T$ & $B_{max}$ 
                      & $T$ & $B_{max}$ & $T$ & $B_{max}$ & $T$ & $B_{max}$ \\
\midrule

\textbf{Openness} 
& 3.1 & -3.5 \scriptsize{(Neu)} & 3.0 & -3.4 \scriptsize{(Neu)} & 2.9 & -3.0 \scriptsize{(Neu)} 
& 3.3 & 3.3 \scriptsize{(Agr)} & 3.2 & 3.4 \scriptsize{(Agr)} & 3.1 & 2.8 \scriptsize{(Agr)} \\
% \addlinespace[0.4em]

\textbf{Conscientiousness} 
& 2.9 & -2.9 \scriptsize{(Neu)} & 2.9 & -2.7 \scriptsize{(Neu)} & 2.9 & 2.6 \scriptsize{(Agr)} 
& 2.2 & -2.0 \scriptsize{(Neu} & 2.3 & 2.0 \scriptsize{(Agr)} & 2.4 & 2.0 \scriptsize{(Agr)} \\
% \addlinespace[0.4em]

\textbf{Extraversion} 
& 3.0 & -3.1 \scriptsize{(Neu)} & 2.6 & -2.4 \scriptsize{(Neu)} & 3.0 & -2.5 \scriptsize{(Neu)} 
& 3.1 & 3.3 \scriptsize{(Agr)} & 3.5 & 3.1 \scriptsize{(Agr)} & 3.3 & 3.0 \scriptsize{(Agr)} \\
% \addlinespace[0.4em]

\textbf{Agreeableness} 
& 3.3 & 2.8 \scriptsize{(Con)} & 3.2 & 2.6 \scriptsize{(Opn)} & \textbf{3.7} & -3.1 \scriptsize{(Neu)} 
& 2.7 & -3.3 \scriptsize{(Neu)} & 3.7 & -3.0 \scriptsize{(Neu)} & 2.8 & -2.3 \scriptsize{(Neu)} \\
% \addlinespace[0.4em]

\textbf{Neuroticism} 
& 3.1 & -3.1 \scriptsize{(Agr)} & 3.1 & -3.2 \scriptsize{(Agr)} & 3.2 & -3.1 \scriptsize{(Agr)} 
& 0.7 & -1.2 \scriptsize{(Ext)} & 0.0 & -1.0 \scriptsize{(Agr)} & 0.1 & -1.1 \scriptsize{(Ext)} \\

\bottomrule
% \caption{\textbf{Trait-Level Steering Contrast under Geometric Constraints.}
% Comparison of target steering contrast ($T$, Intended Target) and maximum cross-trait contrast ($B_{\max}$, Unintended Target) for LLaMA-3-8B and Mistral-8B under baseline (C0), soft-constrained (C4), and hard orthonormal (C5) trait vector constructions. Values report the difference between positively and negatively steered generations (High$-$Low) as measured by judge scores (1--5 scale). Parentheses indicate the trait responsible for $B_{\max}$.}
\end{tabular}
}
\begin{flushleft}
\caption{\textbf{Trait-Level Steering Contrast under Geometric Constraints.}
Comparison of target steering contrast ($T$, Intended Target) and maximum cross-trait bleed ($B_{\max}$, Unintended Target) for LLaMA-3-8B and Mistral-8B under baseline (C0), soft-constrained (C4), and hard orthonormal (C5) trait vector constructions. Values report the difference between positively and negatively steered generations (High$-$Low) as measured by judge scores (1--5 scale). Parentheses indicate the trait responsible for $B_{\max}$.}
\footnotesize{$T$: Diagonal magnitude (Targeted Trait Steering). $B_{max}$: Maximum absolute off-diagonal value.
\textbf{C4 (Soft)} uses $\beta=0.5$. \textbf{C5 (Hard)} uses full symmetric Orthogonalisation. ($-$) sign suggests the opposite nature of the trait effects.}
\label{tab:main_results_trait_resolved}
\end{flushleft}
\end{table*}

% \vspace{-0.1cm}
\section{Evaluation}
% \vspace{-0.45cm} %this deformed the spacing 

Personality steering is evaluated under three controlled settings: \textit{base} (no steering), \textit{positive steering}, and \textit{negative steering}, where the latter two apply trait-specific steering vectors of equal magnitude and opposite polarity. For each geometric condition (C0--C5), all other factors are held constant, including the learned subspace, layer weights, steering intensity, injection point, and decoding configuration. This isolates the effect of geometric constraints on personality vectors, ensuring that observed differences arise solely from vector structure rather than the steering mechanism itself.

Steering effectiveness is assessed using interview-style Big Five Inventory (BFI) questionnaires \cite{wang2024incharacter} consistent with \cite{bhandari2025activation}. For direct comparison, first and second order statistics are reported.
%reporting mean and variance of trait scores across prompts. 
%In addition to target trait shifts, we \emph{explicitly measure cross-trait responses to quantify how steering one trait affects others}, enabling direct analysis of trait interdependence. We perform generation-based evaluation without prompt customisation (neutral prompts), ensuring personality shifts emerge from internal steering alone. 
Beyond target trait shifts, we measure cross-trait responses to quantify inter-trait effects. Evaluation uses neutral prompts, ensuring observed personality changes arise solely from internal steering. We use Gpt-4o-mini as a judge, building upon the literature \cite{jiang2024personallm,frisch2024llm} of using models as judges. Finally, we report fluency scores alongside personality metrics to verify that steering and geometric constraints do not degrade generation quality or general language behaviour.

\section{Results}
We conduct experiments on two instruction-tuned models from different architectural families -- \emph{LLaMA-3-8B-Instruct} and \emph{Ministral-8B-Instruct}. For each model, baseline steering performance is quantified using the difference between positively and negatively steered generations (\emph{high}–\emph{low}), which serves as a reference point to compare how geometric constraints (C1–C5) alter both target-trait control and cross-trait interactions.

\subsection{Geometric Independence of Personality Steering Directions}

To evaluate whether personality steering directions are geometrically independent, we compare the \emph{target steering strength} ($T$, diagonal entries) against the \emph{maximum cross-trait bleed} ($B_{\max}$, largest absolute off-diagonal entry) under different geometric constraints (Table~\ref{tab:main_results_trait_resolved}). Across both LLaMA-3-8B and Mistral-8B, steering a single trait produces non-negligible changes in at least one other trait, with $B_{\max}$ often comparable in magnitude to $T$. This pattern persists under both soft disentanglement (C4) and full symmetric orthonormalisation (C5). Notably, while C5 enforces near-zero pairwise cosine similarity between trait directions in activation space, it does not consistently reduce $B_{\max}$ in generation-level evaluations.

These results indicate that eliminating geometric overlap between steering vectors does not guarantee behavioural independence. Even when trait directions are orthonormal by construction, downstream generations continue to exhibit systematic cross-trait interactions. We therefore \textbf{reject the hypothesis in RQ1}, implying that personality steering directions in large language models are not geometrically independent in a behaviourally meaningful sense.

\begin{figure}[t] 
  \centering\fbox{\includegraphics[width=0.96\columnwidth]{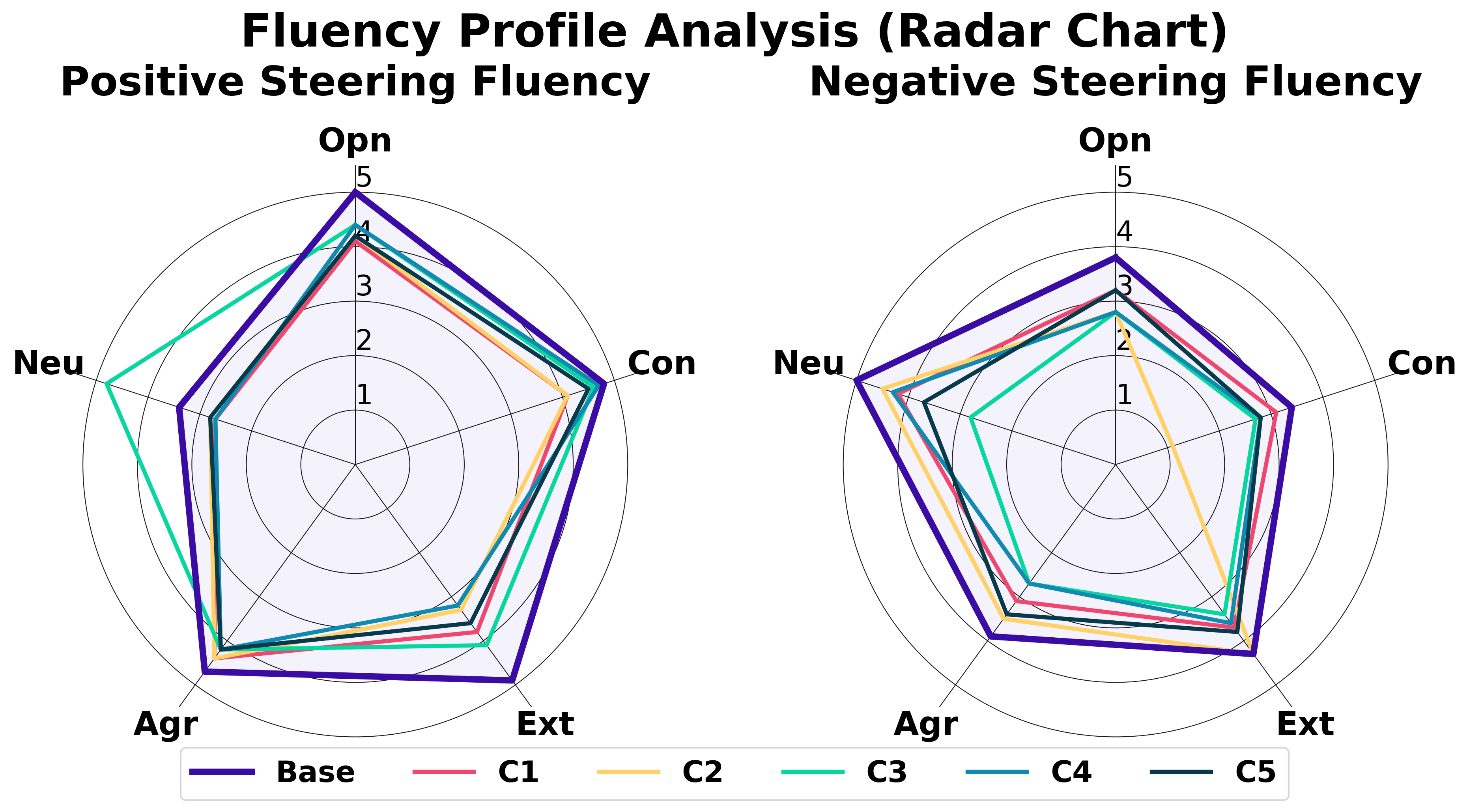}}
  \caption{\textbf{Fluency profile analysis for conditions C0-C5} compared against Base steering. \emph{Fluency degradation for both Positive and Negative steering can be observed across all traits for all the Conditional Methods used}. Although trait shifts that were comparable to the base values, the significant degradation of fluency suggests the need to use orthogonalised vectors carefully for steering purpose. }
  \label{fig:radar_fleuncy_diagram}
  %\vspace{-2mm} % (optional) tighten space below the figure
\end{figure}

Table~\ref{tab:main_results_trait_resolved} reports the steering strength for conditions base(C0), C4 and C5. Additionally, we analyze \emph{fluency} and \emph{variance} scores under identical measurement conditions, using LLaMA as the reference model.

Steering strength remains largely conserved under progressive orthonormalisation (C1--C5)(Table \ref{tab:main_results_trait_resolved}). However, Figure \ref{fig:radar_fleuncy_diagram} demonstrates how fluency scores consistently degrade as geometric constraints are enforced. Comparing to baseline fluency values reported in \cite{bhandari2025activation} (High/Low), Openness drops from $5.0/3.8$ to $4.1/3.2$ (C1), $4.3/2.8$ (C2), and remains around $4.4/2.8$ through C3--C5. Similarly, conscientiousness decreases from $4.8/3.5$ to $4.3/3.1$ in C1 following identical trends to Openness for other conditions, while Extraversion exhibits the largest decline, from $4.9/4.3$ to $3.8/3.6$.

These results indicate that although orthonormalisation preserves directional steering magnitude, it removes shared components necessary for fluent and expressive generation, leading to reduced variance and degraded output quality. Hence, to explain our \textbf{RQ2}: we conclude that explicitly enforcing geometric independence does not improve steering effectiveness and instead introduces a quality–control trade-off.

\subsection{RQ3: Cross-Model Consistency of Trait Dependencies}

To assess whether observed trait dependencies generalise across model families, we compare steering behaviour between \emph{LLaMA-3-8B-Instruct} and \emph{Mistral-8B-Instruct} under identical conditions (C0, C4, C5) using the same extraction, constraint, and evaluation pipeline. Across both models, we observe consistent geometric patterns: several traits exhibit substantial cross-trait bleed even after enforcing geometric constraints. For example, \textbf{Openness} shows high maximum bleed values in both models, with $B_{\max}$ remaining large under hard orthonormalisation (C5), e.g., $\approx 3.0$ in LLaMA-3 and $\approx 2.8$ in Mistral-8B. Similarly, \textbf{Extraversion} and \textbf{Agreeableness} continue to induce strong off-diagonal effects across conditions, indicating that these dependencies are not artifacts of a single model but reflect shared structure in personality representations.

At the same time, we observe clear model-specific modulation in steering responsiveness. Most notably, \textbf{Neuroticism} exhibits strong target steering in LLaMA-3 ($T \approx 3.1$--$3.2$ across C0--C5), whereas Mistral-8B shows low target response ($T \approx 0.0$--$0.7$) under the same conditions, despite comparable geometric treatment. Importantly, this suppression persists even when geometric independence is enforced (C5), suggesting that the absence of behavioural response cannot be attributed solely to vector entanglement. Together, these results indicate that while cross-trait dependencies are largely consistent across model families, their behavioural expression is shaped by model-specific training and alignment constraints rather than geometry alone. A detailed table for all the observations (C1-C5) is provided in Appendix \ref{sec:c0_c5_all_data}.

% Both LLaMA and Mistral exhibit a stable, shared personality subspace with strong inter-trait coupling. However, Mistral shows significantly stronger residual behavioural entanglement than LLaMA, even under strict orthonormal constraints, indicating that geometric disentanglement does not guarantee behavioural independence.

\section{Conclusion}

Given the popularity of steering methods in the literature, we systematically analysed the behaviour of steering vectors under various conditions. Our method considered the Big Five Personality traits, and we investigated whether personality steering directions in large language models are geometrically independent, and how enforcing geometric constraints affects the steering behaviours. Through the analysis across two model families (Llama and Mistral), we show that personality traits are not independent directions in activation space. Even when strong global constraints such as symmetric orthonormalisation are applied, steering one trait consistently induces measurable changes in other unintended traits, indicating persistent cross-trait dependencies.

\section*{Limitations}
This work studies personality steering behaviour using a limited set of large language models and personality datasets, and future work could extend the analysis to a broader range of model families and trait representations. While we focus on Big Five traits and judge-based evaluation, additional datasets and alternative evaluation frameworks may reveal further structure in trait interactions. Our analysis relies on linear geometric constraints; exploring other orthogonalisation or projection methods could provide a more comprehensive understanding of trait disentanglement. Finally, we use LLMs as judges for behavioural assessment, and incorporating human evaluation or complementary metrics is left for future investigation.

\section{Ethical Considerations}
Steering large language models using latent vectors introduces ethical considerations, particularly when such steering is applied in uncontrolled or unsupervised settings. Steering vectors are learned approximations of complex behavioural traits and do not provide transparent or complete representations of the values they encode; as a result, unintended attributes or hidden information may be co-activated during steering. This raises concerns about value misalignment, especially in real-world deployments where subtle behavioural shifts could have downstream social or psychological impacts. Additionally, aggressive or poorly understood steering may bypass safety mechanisms or distort model behaviour in ways that are difficult to detect or reverse. These risks highlight the importance of careful evaluation, interpretability, and constraint-aware steering when modifying model behaviour.

\bibliography{custom}

@inproceedings{jiang2024personallm,
  title={PersonaLLM: Investigating the ability of large language models to express personality traits},
  author={Jiang, Hang and Zhang, Xiajie and Cao, Xubo and Breazeal, Cynthia and Roy, Deb and Kabbara, Jad},
  booktitle={Findings of the association for computational linguistics: NAACL 2024},
  pages={3605--3627},
  year={2024}
}

@article{jiang2023evaluating,
  title={Evaluating and inducing personality in pre-trained language models},
  author={Jiang, Guangyuan and Xu, Manjie and Zhu, Song-Chun and Han, Wenjuan and Zhang, Chi and Zhu, Yixin},
  journal={Advances in Neural Information Processing Systems},
  volume={36},
  pages={10622--10643},
  year={2023}
}

@article{serapio2023personality,
  title={Personality traits in large language models},
  author={Serapio-Garc{\'\i}a, Gregory and Safdari, Mustafa and Crepy, Cl{\'e}ment and Sun, Luning and Fitz, Stephen and Abdulhai, Marwa and Faust, Aleksandra and Matari{\'c}, Maja},
  year={2023}
}

@article{zou2023representation,
  title={Representation engineering: A top-down approach to ai transparency},
  author={Zou, Andy and Phan, Long and Chen, Sarah and Campbell, James and Guo, Phillip and Ren, Richard and Pan, Alexander and Yin, Xuwang and Mazeika, Mantas and Dombrowski, Ann-Kathrin and others},
  journal={arXiv preprint arXiv:2310.01405},
  year={2023}
}

@article{ravfogel2020null,
  title={Null it out: Guarding protected attributes by iterative nullspace projection},
  author={Ravfogel, Shauli and Elazar, Yanai and Gonen, Hila and Twiton, Michael and Goldberg, Yoav},
  journal={arXiv preprint arXiv:2004.07667},
  year={2020}
}

@article{belrose2023leace,
  title={Leace: Perfect linear concept erasure in closed form},
  author={Belrose, Nora and Schneider-Joseph, David and Ravfogel, Shauli and Cotterell, Ryan and Raff, Edward and Biderman, Stella},
  journal={Advances in Neural Information Processing Systems},
  volume={36},
  pages={66044--66063},
  year={2023}
}

@article{elhage2022toy,
  title={Toy models of superposition},
  author={Elhage, Nelson and Hume, Tristan and Olsson, Catherine and Schiefer, Nicholas and Henighan, Tom and Kravec, Shauna and Hatfield-Dodds, Zac and Lasenby, Robert and Drain, Dawn and Chen, Carol and others},
  journal={arXiv preprint arXiv:2209.10652},
  year={2022}
}

@inproceedings{bhandari2025activation,
  title={Activation-Space Personality Steering: Hybrid Layer Selection for Stable Trait Control in {LLMs}},
  author={Bhandari, Pranav and Fay, Nicolas and Selvaganapathy, Sanjeevan and Datta, Amitava and Naseem, Usman and Nasim, Mehwish},
  booktitle={19th Conference of the European Chapter of the Association for Computational Linguistics (EACL'26))},
  year={2026}
}

@inproceedings{li2025big5,
  title={Big5-chat: Shaping llm personalities through training on human-grounded data},
  author={Li, Wenkai and Liu, Jiarui and Liu, Andy and Zhou, Xuhui and Diab, Mona and Sap, Maarten},
  booktitle={Proceedings of the 63rd Annual Meeting of the Association for Computational Linguistics (Volume 1: Long Papers)},
  pages={20434--20471},
  year={2025}
}

@inproceedings{wang2024incharacter,
  title={Incharacter: Evaluating personality fidelity in role-playing agents through psychological interviews},
  author={Wang, Xintao and Xiao, Yunze and Huang, Jen-tse and Yuan, Siyu and Xu, Rui and Guo, Haoran and Tu, Quan and Fei, Yaying and Leng, Ziang and Wang, Wei and others},
  booktitle={Proceedings of the 62nd Annual Meeting of the Association for Computational Linguistics (Volume 1: Long Papers)},
  pages={1840--1873},
  year={2024}
}

@inproceedings{bhandari2025evaluating,
  title={Evaluating personality traits in large language models: Insights from psychological questionnaires},
  author={Bhandari, Pranav and Naseem, Usman and Datta, Amitava and Fay, Nicolas and Nasim, Mehwish},
  booktitle={Companion Proceedings of the ACM on Web Conference 2025},
  pages={868--872},
  year={2025}
}

@article{pellert2024ai,
  title={Ai psychometrics: Assessing the psychological profiles of large language models through psychometric inventories},
  author={Pellert, Max and Lechner, Clemens M and Wagner, Claudia and Rammstedt, Beatrice and Strohmaier, Markus},
  journal={Perspectives on Psychological Science},
  volume={19},
  number={5},
  pages={808--826},
  year={2024},
  publisher={Sage Publications Sage CA: Los Angeles, CA}
}

@article{zhu2024personality,
  title={Personality alignment of large language models},
  author={Zhu, Minjun and Weng, Yixuan and Yang, Linyi and Zhang, Yue},
  journal={arXiv preprint arXiv:2408.11779},
  year={2024}
}

@article{chen2025persona,
  title={Persona vectors: Monitoring and controlling character traits in language models},
  author={Chen, Runjin and Arditi, Andy and Sleight, Henry and Evans, Owain and Lindsey, Jack},
  journal={arXiv preprint arXiv:2507.21509},
  year={2025}
}

@article{yang2024exploring,
  title={Exploring the Personality Traits of LLMs through Latent Features Steering},
  author={Yang, Shu and Zhu, Shenzhe and Liu, Liang and Hu, Lijie and Li, Mengdi and Wang, Di},
  journal={arXiv preprint arXiv:2410.10863},
  year={2024}
}

@inproceedings{sun-etal-2025-personality,
    title = "Personality Vector: Modulating Personality of Large Language Models by Model Merging",
    author = "Sun, Seungjong  and
      Baek, Seo Yeon  and
      Kim, Jang Hyun",
    editor = "Christodoulopoulos, Christos  and
      Chakraborty, Tanmoy  and
      Rose, Carolyn  and
      Peng, Violet",
    booktitle = "Proceedings of the 2025 Conference on Empirical Methods in Natural Language Processing",
    month = nov,
    year = "2025",
    address = "Suzhou, China",
    publisher = "Association for Computational Linguistics",
    url = "https://aclanthology.org/2025.emnlp-main.1253/",
    doi = "10.18653/v1/2025.emnlp-main.1253",
    pages = "24656--24677",
    ISBN = "979-8-89176-332-6"
}

@article{frisch2024llm,
  title={LLM agents in interaction: Measuring personality consistency and linguistic alignment in interacting populations of large language models},
  author={Frisch, Ivar and Giulianelli, Mario},
  journal={arXiv preprint arXiv:2402.02896},
  year={2024}
}

\clearpage
\appendix

\onecolumn
% \newpage
\section{All detailed tables for C0-C5 for trait values}
\label{sec:c0_c5_all_data}

\FloatBarrier

\begin{table*}[!th]
\centering
\caption{\textbf{Condition C1 (Soft Symmetric Whitening).} Cross-trait impact of steering vectors on Llama-3-8B. The rows represent the \textit{Targeted Trait} (steering vector applied), and columns represent the \textit{Measured Trait} (judged output). Values indicate the shift in Likert score from High to Low values(High -- Low). Note the high off-diagonal bleed, particularly between Openness and Extraversion.}
\label{tab:c1_results}
\resizebox{\textwidth}{!}{%
\begin{tabular}{lrrrrr}
\toprule
 & \multicolumn{5}{c}{\textbf{Measured Trait}} \\
\cmidrule(lr){2-6}
\textbf{Targeted Trait} & \textbf{Openness} & \textbf{Conscientiousness} & \textbf{Extraversion} & \textbf{Agreeableness} & \textbf{Neuroticism} \\
\midrule
Openness          & 2.80 &  1.90 &  2.90 &  2.70 & -3.00 \\
Conscientiousness & 1.44 &  3.11 & -1.22 &  2.44 & -2.00 \\
Extraversion      & 2.00 & -0.38 &  2.75 &  1.50 & -2.75 \\
Agreeableness     & 3.11 &  2.56 &  2.67 &  3.67 & -3.22 \\
Neuroticism       & -1.50 & -2.50 & -1.63 & -3.13 &  3.25 \\
\bottomrule
\end{tabular}
}
\end{table*}

\begin{table*}[!th]
\centering
\caption{\textbf{Condition C2 (Greedy Orthogonalisation).}}
\label{tab:c2_results}
\resizebox{\textwidth}{!}{%
\begin{tabular}{lrrrrr}
\toprule
 & \multicolumn{5}{c}{\textbf{Measured Trait}} \\
\cmidrule(lr){2-6}
\textbf{Targeted Trait} & \textbf{Openness} & \textbf{Conscientiousness} & \textbf{Extraversion} & \textbf{Agreeableness} & \textbf{Neuroticism} \\
\midrule
Openness          & 2.50 & 2.20 & 2.30 & 1.80 &  -2.50 \\
Conscientiousness &  0.11 & -3.00 &  2.33 &  0.22 & -0.11 \\
Extraversion      & -2.00 &  0.63 & -3.13 &  0.88 &  2.88 \\
Agreeableness     & 1.89 & 0.67 & 0.67 & 2.67 &  -1.00 \\
Neuroticism       & -1.63 & -2.88 & -2.13 & -3.38 &  3.63 \\
\bottomrule
\end{tabular}
}
\end{table*}

\begin{table*}[!th]
\centering
\caption{\textbf{Condition C3 (Selective Orthogonalisation).}}
\label{tab:c3_results}
\resizebox{\textwidth}{!}{%
\begin{tabular}{lrrrrr}
\toprule
 & \multicolumn{5}{c}{\textbf{Measured Trait}} \\
\cmidrule(lr){2-6}
\textbf{Targeted Trait} & \textbf{Openness} & \textbf{Conscientiousness} & \textbf{Extraversion} & \textbf{Agreeableness} & \textbf{Neuroticism} \\
\midrule
Openness          &  3.10 &  2.50 &  2.90 &  3.30 & -3.40 \\
Conscientiousness &  1.56 &  2.89 & -1.22 &  2.44 & -2.75 \\
Extraversion      &  2.00 & -1.75 &  2.75 &  1.38 &  0.13 \\
Agreeableness     & 2.56 & 2.67 & 1.67 & 3.22 &  -3.00 \\
Neuroticism       & -1.50 & -2.88 & -0.50 & -3.63 &  2.75 \\
\bottomrule
\end{tabular}
}
\end{table*}

% \section{}
\begin{table*}[!th]
\centering
\caption{\textbf{Condition C4 (Soft Greedy Projection, $\beta=0.5$).}}
\label{tab:c4_results}
\resizebox{\textwidth}{!}{%
\begin{tabular}{lrrrrr}
\toprule
 & \multicolumn{5}{c}{\textbf{Measured Trait}} \\
\cmidrule(lr){2-6}
\textbf{Targeted Trait} & \textbf{Openness} & \textbf{Conscientiousness} & \textbf{Extraversion} & \textbf{Agreeableness} & \textbf{Neuroticism} \\
\midrule
Openness          &  3.00 &  2.60 &  2.90 &  3.30 & -3.40 \\
Conscientiousness &  1.78 &  2.88 & -1.22 &  2.44 & -2.75 \\
Extraversion      &  2.00 & -0.88 &  2.63 &  1.38 & -2.38 \\
Agreeableness     & 2.56 & 2.33 & 1.67 & 3.22 &  -2.33 \\
Neuroticism       & -1.38 & -2.63 & -2.25 & -3.25 &  3.13 \\
\bottomrule
\end{tabular}
}
\end{table*}

\begin{table*}[!th]
\centering
\caption{\textbf{Condition C5 (Hard Symmetric Orthonormalisation).}}
\label{tab:c5_results}
\resizebox{\textwidth}{!}{%
\begin{tabular}{lrrrrr}
\toprule
 & \multicolumn{5}{c}{\textbf{Measured Trait}} \\
\cmidrule(lr){2-6}
\textbf{Targeted Trait} & \textbf{Openness} & \textbf{Conscientiousness} & \textbf{Extraversion} & \textbf{Agreeableness} & \textbf{Neuroticism} \\
\midrule
Openness          &  2.90 &  1.90 &  3.00 &  2.70 & -3.00 \\
Conscientiousness &  1.44 &  2.89 & -1.11 &  2.56 & -2.00 \\
Extraversion      &  1.88 & -0.38 &  3.00 &  1.25 & -2.50 \\
Agreeableness     &  2.89 &  2.78 &  2.44 &  3.67 & -3.11 \\
Neuroticism       & -1.38 & -2.50 & -1.63 & -3.13 &  3.25 \\
\bottomrule
\end{tabular}
}
\end{table*}

\begin{table*}[!th]
\centering
\caption{\textbf{Condition C1 (Soft Symmetric Whitening) on Mistral-8B.}}
\label{tab:c1_mistral_results}
\resizebox{\textwidth}{!}{%
\begin{tabular}{lrrrrr}
\toprule
 & \multicolumn{5}{c}{\textbf{Measured Trait}} \\
\cmidrule(lr){2-6}
\textbf{Targeted Trait} & \textbf{Openness} & \textbf{Conscientiousness} & \textbf{Extraversion} & \textbf{Agreeableness} & \textbf{Neuroticism} \\
\midrule
Openness          &  3.10 &  2.40 &  2.80 &  2.80 & -2.50 \\
Conscientiousness &  1.11 &  2.44 &  0.44 &  2.00 & -1.89 \\
Extraversion      &  2.38 &  2.00 &  3.38 &  2.75 & -1.75 \\
Agreeableness     &  2.00 &  0.56 &  2.22 &  2.78 & -2.22 \\
Neuroticism       & -0.50 & -1.13 & -1.13 & -0.75 & -0.13 \\
\bottomrule
\end{tabular}
}
\end{table*}

\begin{table*}[!th]
\centering
\caption{\textbf{Condition C2 (Greedy Orthogonalisation) on Mistral-8B.}}
\label{tab:c2_mistral_results}
\resizebox{\textwidth}{!}{%
\begin{tabular}{lrrrrr}
\toprule
 & \multicolumn{5}{c}{\textbf{Measured Trait}} \\
\cmidrule(lr){2-6}
\textbf{Targeted Trait} & \textbf{Openness} & \textbf{Conscientiousness} & \textbf{Extraversion} & \textbf{Agreeableness} & \textbf{Neuroticism} \\
\midrule
Openness          & 2.50 & 2.10 & 2.40 & 2.10 &  -0.40 \\
Conscientiousness &  0.89 &  2.22 &  0.11 &  2.22 & -1.67 \\
Extraversion      &  2.13 &  1.75 &  3.13 &  2.88 & -2.25 \\
Agreeableness     &  1.56 &  0.44 &  2.00 &  2.89 & -2.22 \\
Neuroticism       & -1.13 & -1.50 & -1.25 & -1.00 & -1.00 \\
\bottomrule
\end{tabular}
}
\end{table*}

\begin{table*}[!th]
\centering
\caption{\textbf{Condition C3 (Selective Orthogonalisation) on Mistral-8B.}}
\label{tab:c3_mistral_results}
\resizebox{\textwidth}{!}{%
\begin{tabular}{lrrrrr}
\toprule
 & \multicolumn{5}{c}{\textbf{Measured Trait}} \\
\cmidrule(lr){2-6}
\textbf{Targeted Trait} & \textbf{Openness} & \textbf{Conscientiousness} & \textbf{Extraversion} & \textbf{Agreeableness} & \textbf{Neuroticism} \\
\midrule
Openness          &  3.30 &  2.60 &  3.00 &  3.30 & -2.30 \\
Conscientiousness &  1.11 &  2.22 &  0.89 &  2.00 & -1.89 \\
Extraversion      &  2.63 &  1.50 &  3.63 &  3.25 & -2.13 \\
Agreeableness     &  2.11 &  1.33 &  2.11 &  3.33 & -2.89 \\
Neuroticism       & -0.25 & -1.25 & -0.38 &  0.13 & -0.25 \\
\bottomrule
\end{tabular}
}
\end{table*}

\begin{table*}[!th]
\centering
\caption{\textbf{Condition C4 (Soft Greedy Projection, $\beta=0.5$) on Mistral-8B.}}
\label{tab:c4_mistral_results}
\resizebox{\textwidth}{!}{%
\begin{tabular}{lrrrrr}
\toprule
 & \multicolumn{5}{c}{\textbf{Measured Trait}} \\
\cmidrule(lr){2-6}
\textbf{Targeted Trait} & \textbf{Openness} & \textbf{Conscientiousness} & \textbf{Extraversion} & \textbf{Agreeableness} & \textbf{Neuroticism} \\
\midrule
Openness          &  3.20 &  2.40 &  3.10 &  3.40 & -2.40 \\
Conscientiousness &  1.00 &  2.33 &  1.00 &  2.00 & -1.67 \\
Extraversion      &  2.50 &  1.88 &  3.50 &  3.13 & -2.00 \\
Agreeableness     &  2.11 &  1.89 &  2.44 &  3.67 & -3.00 \\
Neuroticism       & -0.50 & -0.88 & -0.50 & -1.00 &  0.00 \\
\bottomrule
\end{tabular}
}
\end{table*}
% \vspace{-6cm}

\begin{table*}[!th]
\centering
\caption{\textbf{Condition C5 (Hard Symmetric Orthonormalisation) on Mistral-8B.}}
\label{tab:c5_mistral_results}
\resizebox{\textwidth}{!}{%
\begin{tabular}{lrrrrr}
\toprule
 & \multicolumn{5}{c}{\textbf{Measured Trait}} \\
\cmidrule(lr){2-6}
\textbf{Targeted Trait} & \textbf{Openness} & \textbf{Conscientiousness} & \textbf{Extraversion} & \textbf{Agreeableness} & \textbf{Neuroticism} \\
\midrule
Openness          &  3.10 &  2.30 &  2.70 &  2.80 & -2.50 \\
Conscientiousness &  1.11 &  2.44 &  0.67 &  2.00 & -1.56 \\
Extraversion      &  2.38 &  2.13 &  3.25 &  3.00 & -1.63 \\
Agreeableness     &  2.00 &  0.56 &  2.22 &  2.78 & -2.33 \\
Neuroticism       & -0.50 & -0.88 & -1.13 & -0.50 & -0.13 \\
\bottomrule
\end{tabular}
}
\end{table*}

\begin{table*}[!th]
\centering
\small
\caption{
Diagnostics of progressive orthonormalization constraints (C1--C5) applied to personality steering directions in LLaMA-3-8B.
For each constraint, we report the achieved geometric independence (maximum absolute off-diagonal cosine similarity between trait directions), the range of signal retention relative to the unconstrained baseline, and a brief qualitative summary.
C1 and C5 enforce global, order-independent orthonormality and achieve near-zero cosine overlap, but uniformly attenuate trait signal.
C2 also enforces strict orthonormality, but its greedy, order-dependent construction leads to severe semantic degradation.
C3 and C4 relax hard orthogonality by selectively or softly removing projections, preserving substantially more trait signal at the cost of residual geometric entanglement, with C4 exhibiting the best semantic–geometry trade-off.
}
\label{tab:ortho_diagnostics}
\resizebox{\textwidth}{!}{%
\begin{tabular}{c l c c c}
\toprule
\textbf{C} & \textbf{Method} & \textbf{Geom. Independence} & \textbf{Signal Retention} & \textbf{Key Note} \\
\midrule
C1 & Global Gram whitening 
& $\max|\cos| < 10^{-8}$ 
& $0.83$--$0.94$ 
& Perfect ortho; mild attenuation \\
C2 & Strict QR (order-dependent) 
& $\max|\cos| < 10^{-8}$ 
& $-1.00$--$0.63$ 
& Order effects destroy trait semantics \\
C3 & Selective removal ($\tau{=}0.5$) 
& $\max|\cos|=0.466$ 
& $0.63$--$1.00$ 
& Partial decorrelation; E most affected \\
C4 & Soft removal ($\beta{=}0.5,\tau{=}0.5$) 
& $\max|\cos|=0.527$ 
& $0.85$--$1.00$ 
& Best trade-off; semantics largely preserved \\
C5 & Hard orthonormal (global) 
& $\max|\cos| < 10^{-8}$ 
& $0.83$--$0.94$ 
& Perfect ortho; signal uniformly reduced \\
\bottomrule
\end{tabular}
}
\end{table*}

\end{document}